\documentclass[11pt]{article}
\pdfoutput=1

\usepackage[]{acl}

\usepackage{times}
\usepackage{latexsym}
\usepackage[T1]{fontenc}
\usepackage[utf8]{inputenc}
\usepackage{microtype}
\usepackage{inconsolata}
\usepackage{multirow}
\usepackage{float}
\usepackage{graphicx}

\title{Automatic Labelling with Open-source LLMs using Dynamic Label Schema Integration}

\author{
\textbf{Thomas Walshe\thanks{Work carried out during an internship at JP Morgan Chase Co.}}$^{2}$~~~
\textbf{Sae Young Moon}$^1$~~~
\textbf{Chunyang Xiao}$^1$~~~ \\
\textbf{Yawwani Gunawardana}$^{1}$~~~ 
\textbf{Fran Silavong}$^{1}$~~~\\
$^1$J.P. Morgan Chase \\
\texttt{}
$^2$Snorkel AI\\
\texttt{}
\texttt{thomas.walshe@snorkel.ai}~~~
\texttt{saeyoung.moon@jpmchase.com} \\
\texttt{chunyang.xiao@jpmchase.com}~~~
\texttt{yawwani.p.gunawardana@jpmorgan.com}~~~ \\
\texttt{fran.silavong@jpmchase.com}
}

\begin{document}
\maketitle
\begin{abstract}
%Acquiring sufficient quantities of accurate labelled training data remains a costly task in machine learning projects, often requiring significant human input to meet quantity and quality requirements. We propose a novel strategy for applying Large Language Models (LLMs) to the task of data annotation where labelled data is initially limited. Through a multi-agent framework, we first introduce a self-improvement mechanism that generates and refines annotation guidelines for a given label schema. Over multiple refinement iterations, the guidelines are automatically updated to for account any shortcomings or gaps in the instructions when used in LLM-based annotation and evaluated against a small dataset of ground-truth data. Further, as an alternative to majority voting or meta-reasoning, we demonstrate that multiple versions of guidelines can be used in a weak supervision scenario to improve the accuracy of assigned labels. 
Acquiring labelled training data remains a costly task in real world machine learning projects to meet quantity and quality requirements. Recently Large Language Models (LLMs), notably GPT-4, have shown great promises in labelling data with high accuracy. However, privacy and cost concerns prevent the ubiquitous use of GPT-4. In this work, we explore effectively leveraging open-source models for automatic labelling. We identify integrating label schema as a promising technology but found that naively using the label description for classification leads to poor performance on high cardinality tasks. To address this, we propose Retrieval Augmented Classification (RAC) for which LLM performs inferences for one label at a time using corresponding label schema; we start with the most related label and iterates until a label is chosen by the LLM. We show that our method, which dynamically integrates label description, leads to performance improvements in labelling tasks. We further show that by focusing only on the most promising labels, RAC can trade off between label quality and coverage - a property we leverage to automatically label our internal datasets. 

\end{abstract}

\section{Introduction}
\label{section:introduction}
%While model-centric research has long been the focus of academic studies~\citep{kadhim2019survey}, industry 

Real-life applications must go beyond model architecture choices and consider all stages of the ML development lifecycle~\citep{de2019understanding}. In the case of supervised learning, central to this process is a requirement to possess or create a high quality labelled dataset that reflects the goals of the business, often facilitated through an iterative annotation process~\citep{polyzotis2018data, tseng2020best}. Despite recent popularity of platforms for crowdsourced labelling, such as Mechanical Turk (MTurk)~\citep{sorokin2008utility}, and active studies in this domain, annotation remain costly in terms of both time and money~\citep{wang-etal-2021-want-reduce}. 

%The creation of a new dataset is often facilitated through an iterative process during which functional requirements are outlined, label schemas are defined (flat or hierarchical structures defining and describe the label space), annotation guidelines are created, data is annotated, and the quality of the annotations is assessed~\citep{tseng2020best}. Issues at any stage may necessitate costly modifications to the schema, clarifications to be added to the guidelines, and data to be re-tagged to reflect the changes or meet quality requirements. 

Recently, Large Language Models (LLMs)~\citep{rae2021scaling, thoppilan2022lamda, chowdhery2022palm, touvron2023llama} have emerged as generalist models, and have demonstrated promising performance in zero-shot~\citep{meng2022generating, kojima2022large} and few-shot~\citep{brown2020language, schick2020s} tasks. In the context of data annotation, there has been significant interest in leveraging these powerful generalist models as cost-effective data labelers~\citep{gilardi2023chatgpt, fonseca2023can}, augmenters~\citep{moller2023prompt}, and generators~\citep{lee2023making, wang2023let}.
However, many of these works have focused on the use of externally-hosted LLMs that are made accessible through public-facing APIs and billing that operates on a per-token basis. Certain scenarios, particularly those in which data privacy is paramount, necessitate that models are hosted in-house. In this work, we study whether open-source LLMs, hosted internally, can generate high quality labels for classification tasks. Furthermore, we conduct labelling in a zero-shot scenario to minimise the amount of human effort required for annotation, and use 7B models to account for resource provision constraints. 

We investigate label quality by benchmarking on classification tasks of various levels of difficulty: simple ones usually involve a few well separated categories (low cardinality) and complex ones, such as our internal dataset, involve more nuanced categories (high cardinality, see Figure~\ref{figure:rac_diagram} for illustration) . In Section~\ref{section:preliminary_experiments:experiment_results}, we conduct preliminary experiments that demonstrate the benefit to labelling accuracy that comes with providing more detailed label schema information in the prompt. However, we find that classification performance degrades with high cardinality datasets if all label schema information is provided at once, probably because LLMs can't leverage all information precisely for inference~\cite{liu2023lost}. To mitigate this issue, we cast the problem of multi-class classification as a series of binary ones~\citep{allwein00reducing}, and further propose Retrieval Augmented Classification (RAC) as a solution. RAC orders the labels in decreasing order of semantic relatedness between queries (text to be labelled) and categories, which are incorporated into a binary classification prompt. RAC effectively leverages label schema and generates reasonably high quality data labels. Furthermore, the framework naturally allows abstaining by only perform binary predictions on the first retrieved labels. We show empirically that this can let the model to produce most of the labels (>70\%) but with much higher label accuracy on public benchmarks. We present our industry use case and discuss in further details the balance between label quality and dataset coverage for our internal dataset which we leverage RAC to label automatically.  

%We measure our progress by first constructing a label generation upperbound by finetuning domain classifiers and then measure the gap between automatic labelling and these domain classifiers. Even our best open-source LLMs show promises but have a relatively large gap compared to domain classifiers. To bridge the gap, we further benchmarked the performance when we include COT [CITATION], self consistency [CITATION] and label schema where self consistency and label schema improves slightly the results. An issue we notice is that while LLMs can confidently give correct labels for many data, there are also data which is beyond its reach.   

%In this work, we look to use LLMs for zero-shot classification, specifically focusing on (relatively) small 7-billion parameter models, by utilising the information available within the schemas and guidelines typically available at annotation time. To alleviate the need to manually refine annotation guidelines based on feedback from human-in-the-loop reviewers, we also propose multi-agent framework using LLMs for guideline improvement. 

\section{Preliminary Studies}
\label{section:preliminary_experiments}

\subsection{Leveraging label description for classification}
\label{section:preliminary_experiments:enriching_label_schema}

In alignment with existing works~\cite{fonseca2023can}, we evaluate the effect of leveraging label description to improve zero-shot classification with LLM. A label schema thus contains both the label name (L) and description (D):

\begin{itemize}
    \item Label Name (L) - Approximately five words that provide a concise name for a given category.
    \item Label Description (D) - Approximately 50 words that describe in greater detail the semantics of a given category. For intent classification this could include descriptions of user actions or objectives~\citep{shah2023using}.
\end{itemize}

For internal experiments, we receive well-defined label names and descriptions in our case manually created by Subject Matter Experts (SMEs), allowing for a good understanding of each category. However, for public datasets, similar annotation materials are not commonly provided. To investigate into the effect of label description, we use Mistral-7B to generate the label description by giving in prompt a small quantity of training examples. We use prompts based on those of~\citet{shah2023using} with exact prompts shown in  Figure~\ref{figure:prompt_description_generation} in the Appendix.\footnote{For dataset like Banking77, we also found label names sometimes not reflecting well the category. Due to this fact, we generate a new label using prompt shown in Figure~\ref{figure:prompt_label_generation} to replace the original label.}

%Meta-reasoning~\citep{yoran2023answering} over multiple Chains-of-Thought~\citep{wei2022chain} is found to improve the subjective quality of the generated label names and descriptions, however we leave the systematic evaluation of label information for future work. For each category in each dataset, we generate 5 permutations of the written label names and description.

\subsection{Experiment settings}
\label{section:preliminary_experiments:experiment_settings}
We choose two open-sourced LLMs, namely Mistral-7B~\citep{jiang2023mistral} and Llama-7B \cite{touvron2023llama}, to evaluate whether their performance is sufficient for industry-grade data labelling. To identify optimal prompting approaches for labelling, we vary the label schema component included in the prompt and examine the effect of Chain-of-Thought prompting (COT) \cite{wei2023chainofthought}. The specific prompts are outlined in the Appendix, non-COT prompt in Figure ~\ref{figure:prompt_label_classification_no_cot} and COT prompt in Figure~\ref{figure:prompt_label_classification_cot}.

We use AGNews~\citep{zhang2015character}, DBpedia~\citep{lehmann2015dbpedia}, Amazon intents~\citep{bastianelli2020slurp, fitzgerald2022massive}, and Banking77~\citep{casanueva2020efficient} for our preliminary study. To compare across experiment configurations, we construct a per-class binary classification problem for each dataset where for each class, a balanced set of positives and negatives are sampled from the withheld test splits. For the Banking77 and Amazon datasets, we sample the maximum possible per class, for AGNews and DBpedia we sample 512 positives and 512 negatives for each class. For efficient offline batch inference during the experiments, vLLM is used as a backend~\citep{kwon2023efficient}. We report macro-averaged accuracy.

\subsection{LLM-based binary classification results}
\label{section:preliminary_experiments:experiment_results}

\begin{table}[t!]
  \centering
\begin{tabular}{lllllll}
 & \multicolumn{2}{c}{L} & \multicolumn{2}{c}{L + D} \\
 Dataset &   No CoT & CoT &   No CoT & CoT  \\
 \hline
  AGNews   & 76.2 & 80.0 & 83.4 & \textbf{84.4}      \\
 Banking77 & 81.6 & 79.2 & \textbf{88.8}& 83.7 \\
 DBpedia   & 78.0 & 83.1 & 84.6 & \textbf{88.8}  \\
 Amazon    & 85.5 & 86.0 & 88.9 & \textbf{89.2}  \\
\end{tabular}
  \caption{Macro-averaged zero-shot classification accuracy across datasets, using Mistral 7B with varied prompts. (L) uses only label names while (L+D) provides the full label schema including label names and descriptions.}
  \label{table:llm_classifiers}
\end{table}
From our preliminary results, we observe that Mistral 7B significantly outperforms Llama 7B on this binary classification task (results are shown in Table~\ref{table:appendix:llama_results} in the Appendix), thus we use Mistral 7B as our LLM for the rest of our experiments in the paper.\footnote{The fact that we generate label descriptions from Mistral 7B might bias this comparison, for which we leave the thorough investigation for future work.}

Results are shown in Table~\ref{table:llm_classifiers}. Firstly, we observe that including label description always improves labelling performance for all settings we tested (8 in total with 4 datasets and both COT and non-COT promptings). Secondly, although the use of COT often results in performance gains, this is not consistent across all datasets. 

We also explore several self consistency methods with preliminary results shown in Appendix~\ref{section:appendix:self_consistency_experiments}. We observe in our case that majority voting outperforms other more sophisticated methods such as weak supervision and meta-reasoning.

\noindent\textbf{Following experimental setups} Based on the preliminary results, we investigate further incorporating label description into multiclass classification problems. We choose Mistral-7B as our LLM. When we use LLM to make a binary classification, we only output a prediction when no-COT and COT approaches output the same label. We only use majority voting as our self consistency method.

\section{Retrieval Augmented Classification}

%\subsection{RAC Model}
\label{section:methodology:rac}
Given the potential of label description to improve performance for automatic labelling, we first conduct experiments where we incorporate all label names and descriptions into an LLM as a single prompt, letting LLM act as multi-class classifier. We find a significant degradation in performance in the challenging, high cardinality setting. For example, on Banking77, a consumer banking customer intent classification dataset with 77 categories, we observe a 4.0 F1 with Mistral-7B (macro-averaged). We hypothesise that, despite the supported large context length, the LLM is unable to handle all the label and their information effectively in the prompt, which is consistent with what literature reports as `lost in the middle'~\cite{liu2023lost}. 

\begin{figure*}[t!]
    \centering
    \includegraphics[width=0.65\textwidth]{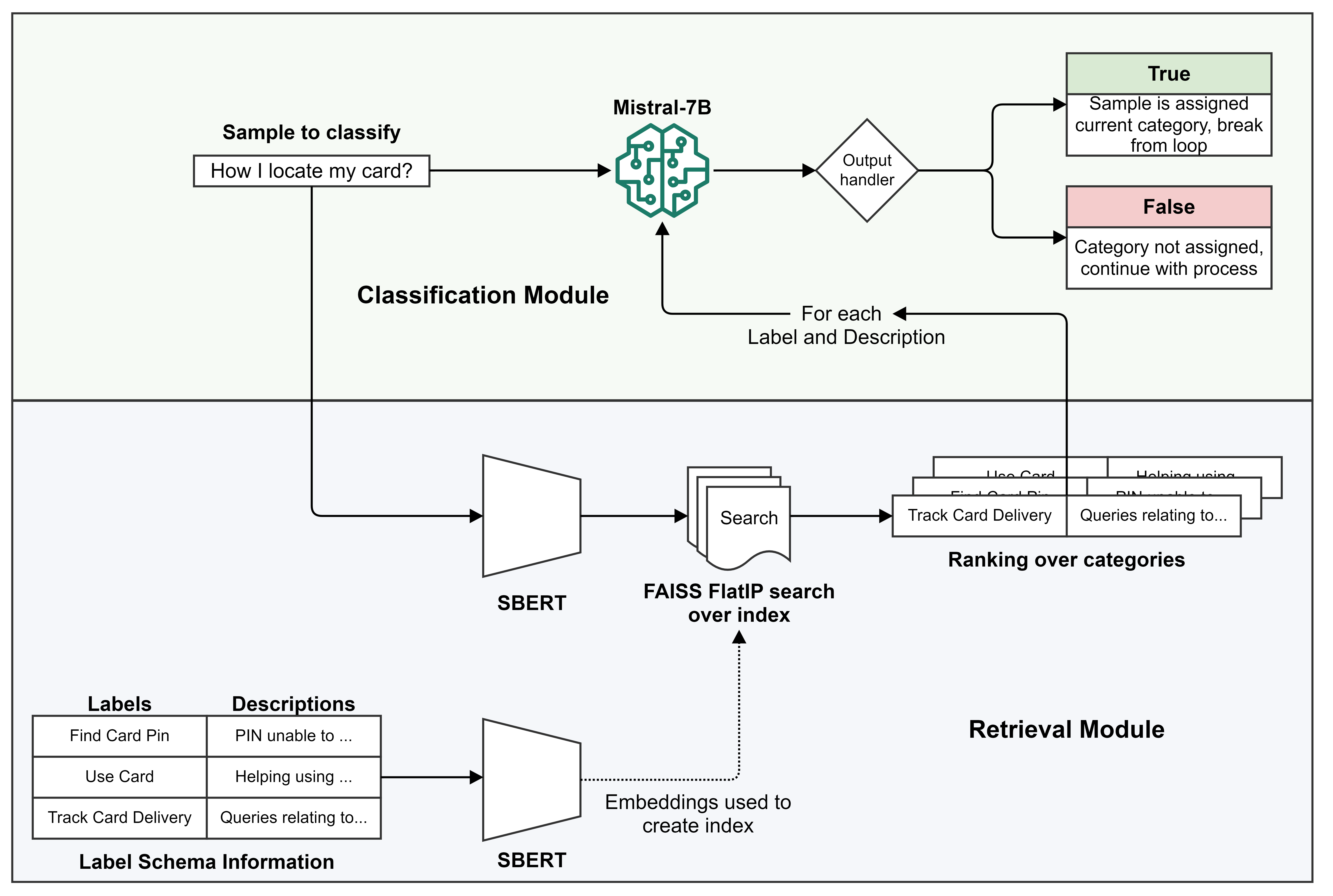}
    \caption{Retrieval Augmented Classification (RAC) during inference: the (bottom) retrieval part retrieved most relevant category and its corresponding description; (top) LLM classification leverages these information and performs binary classification iteratively.}
    \label{figure:rac_diagram}
\end{figure*}

To mitigate the long prompt issue, we leverage the well established idea to convert the multi-class classification problem into a series of binary ones~\citep{allwein00reducing}, allowing the LLM to focus on label schema information for a single category at a time. The optimal configuration for conducting zero-shot binary classification using LLM was chosen through our preliminary studies in Section \ref{section:preliminary_experiments}. Furthermore, drawing inspiration from existing approaches using Retrieval Augmented Classification (RAC)~\citep{long2022retrieval, iscen2023improving}, we propose to incorporate a retrieval component so that binary classification is done iteratively starting from the most promising labels. The process is illustrated in Figure ~\ref{figure:rac_diagram}:
\begin{enumerate}
    \item Search index construction - Prior to label generation (i.e. inference step), the label description corresponding to each category is embedded and a search index created and stored, enabling a nearest neighbour search to the most semantically related category given a query. At inference time, a ranked retrieved list is produced as input to the LLM, incorporating label name and description as part of the prompts.
    \item Iterative classification - For each category in the ranked list, and including the sample to be annotated, a zero-shot prompt is formed using the setup described in Section~\ref{section:preliminary_experiments} for LLM to perform binary classification. If the result returns True (i.e., the LLM believes the sample belongs to a given category), the sample is assigned the label of the current category and no further classification is performed (early stopping). The next category is considered if LLM returns False.
\end{enumerate}

Despite the algorithm being simple and intuitive, it has several interesting properties that we highlight below: \newline

\noindent\textbf{Further usage of label description} In our preliminary study,  Section~\ref{section:preliminary_experiments:experiment_results}, we show that label description helps LLMs to perform better in its binary decision to accept the label or not. The label description in RAC is also used in the retrieval component, which we believe should enhance the retriever performance. Retriever performance enhancement not only improves the label accuracy by feeding most relevant labels first, but also improves the system latency as the system can accept a label earlier. We show the retriever performance in this section while leaving the latency discussion later in our ablations studies. 

We benchmark the retrieval performance across each dataset with an off-the-shelf sentence transformer (all-mpnet-base-v2) without considering finetuning~\cite{gao2021simcse,tunstall2022efficient} and quantify the performance by measuring the Mean Reciprocal Rank (MRR) of categories against the queries. A higher MRR@$1$ indicates that the correct category is more likely to be ranked first.\footnote{We compute MRR for other $n$ for different datasets and observe consistent trends.} The results are shown in Table~\ref{table:rac_mrr_small}. Retriever performance is found to be maximised when each category is represented by the written labels and descriptions, showing that label description indeed improves the retrieval performance.

\begin{table}[h]
\centering
\begin{tabular}{lll}
 L & D & L + D\\
 \hline
 0.635 & 0.679 & \textbf{0.691}\\
\end{tabular}
\caption{MRR@1 of the retriever averaged over each dataset. Categories are represented using the label (L), description (D), or a concatenation of both (L+D).}
%Individual labels etc. are embedded and the final MRR is averaged over the 5 permutations of the label schema for each category. This is compared to averaging the embedding representations of all 5 permutations prior to constructing the retriever.}
  \label{table:rac_mrr_small}
\end{table}

\noindent\textbf{Truncated RAC} One approach to use RAC is to iterate through the labels one by one and stop only when one label is accepted (or the categories exhaust). We call this approach \textbf{Full RAC} since the system tries to assign a label for every data point, thus maximising coverage. Another approach we consider in this work called \textbf{Truncated RAC} only iterates the first $n$ labels. In this situation, if the model rejects the $n$ label proposals, it does not output a label and thus abstains from labelling. However, because only high ranked labels are considered and outputted by the LLM, the system arguably only outputs confident predictions. Therefore, Truncated RAC has the potential to balance between coverage and precision to suit different business use cases. We study this question in more details for our public and internal experiment result sections.

\section{Results on public datasets}
\label{section:public_results_and_discussion}

\begin{table*}[t!]
  \centering
\begin{tabular}{lllllllllll}
 & \multicolumn{2}{c}{Full (L)} & \multicolumn{2}{c}{Full (L + D)} & \multicolumn{3}{c}{Truncated (L)} & \multicolumn{3}{c}{Truncated (L + D)}\\
 Dataset &   Micro & Macro &  Micro & Macro & Micro & Macro & Cov. & Micro & Macro & Cov. \\
 \hline
  AGNews    & 67.1 & 67.5 & 69.3 & 69.7 & 83.5 & 82.9 & 67.4 & 86.9 & 87.0 & 71.5  \\
  Banking77 & 56.2 & 55.6 & 66.8 & 68.3 & 61.6 & 59.4 & 90.3 & 74.7 & 73.4 & 88.7 \\ 

 DBpedia    & 23.2 & 25.2 & 43.8 & 51.2 & 90.3 & 82.9 & 18.2 & 96.0 & 94.3 & 38.8  \\
 Amazon     & 42.0 & 47.4 & 53.6 & 57.4 & 64.1 & 62.1 & 63.3 & 73.7 & 68.5 & 70.0 
\end{tabular}

  \caption{Multi-class F1 when using the RAC pipeline across all datasets.}
  \label{table:rac_f1}
\end{table*}

\subsection{Datasets}
\label{section:methodology:datasets}
We evaluate our proposed solution on four public datasets for text classification, ranging in difficulty from low to high cardinality: AGNews (4 categories)~\citep{zhang2015character}, DBpedia (14 categories)~\citep{lehmann2015dbpedia}, Amazon intents (60 categories)~\citep{bastianelli2020slurp, fitzgerald2022massive}, and Banking77 (77 categories)~\citep{casanueva2020efficient}. 

For Amazon, Banking77 and DBpedia, we run Truncated RAC with a maximum number of 5 steps while we run only 2 steps for AGNews since it only contains 4 classes. We also report the coverage for Truncated RAC (Cov., the percentage of the data that is labelled within the limited number of steps). We report micro-F1 and macro-F1 for all experiments. Unlabelled samples are not included in F1 calculations.
%Across all datasets we see consistent increases to the Mean Reciprocal Rank (MRR) when the averaged representation of the 5 label schema permutations is used in the retriever. Except for Banking77, the highest MRR is achieved using a concatenation of category labels and descriptions. Higher MRR values result in greater throughput during offline batch inference as sequential classification is more likely to terminate after fewer steps. 

\subsection{Full and Truncated RAC}
The results in Table~\ref{table:rac_f1} show the results of full and Truncated RAC across each of the public datasets. We observe that 1) Integrating label description consistently improves labelling quality. Across all datasets and RAC settings (i.e. Full and Truncated) (in total 8 settings), we observe improvement when we integrate label description into RAC. The improvement gap ranges from small (around 2\% for AGNews Full setting) to large (around 20\% for DBpedia Full setting). The coverage for the Truncated setting overall improves for 3 datasets among 4, which is consistent with our Table~\ref{table:rac_mrr_small} that label description improves the coverage at the retrieval stage. 2). Truncated RAC effectively trades off between label quality and coverage. By comparing Full (L+D) and Truncated (L+D), Truncated settings improve F1 scores about 20 absolute points on both AGNews and Amazon, while covering around 70\% of the cases.

We analyse the errors made by Truncated RAC. First, we observe that some categories are overly broad. For example, AGNews dataset contains a \textit{World News} category leading to many samples being erroneously labelled. Second, nuanced categorization impacts the LLM performance. For example, in Banking77, the categories \textit{Beneficiary not allowed}, \textit{Transfer failed}, and \textit{Transfer not received by recipient} all discuss bank transfer failures. %This is also seen in the Amazon dataset in categories such as \textit{Datetime query} (e.g., `what time is it in England?') and \textit{Datetime conversion} (e.g., `I need the time zone from London England instead of central time zone').

%Consistent with earlier observations, including the descriptions into the prompt improves labelling accuracy across all datasets. We also observe the benefit to accuracy that comes with truncated RAC, albeit at the cost of label coverage.

\subsection{Analysis}
\subsubsection{Latency investigation}
With the Banking77 dataset, we show in Table~\ref{table:banking77_comparison} our latency study for different RAC configurations. The baseline `All info in prompt' performs poorly despite its fast latency, and the random retriever takes significant amount of time to accept a label. Full RAC improves over random retriever by ranking the label candidates. Truncated RAC further improves F1 and matches almost `All info in prompt' inference time; as a trade-off, it does not cover all instances (88.7\% shown in Table~\ref{table:rac_f1}).

\begin{table}[t!]
\centering
\begin{tabular}{lll}
 Configuration &  Time (s) &  F1\\
 \hline
 All info in prompt &  3.0 &  4.0\\
 Random retriever & 42.5 & 58.4\\
 Full RAC & 18.8 & 68.3\\
 Truncated RAC & 3.8 & 73.4 
\end{tabular}
\caption{Comparison of the performance of each configuration on the Banking77 dataset showing the average time to label a sample and the macro-averaged F1. `All info in prompt' includes the entire label schema. `Random retriever' removes the semantic ranking over the categories.}
  \label{table:banking77_comparison}
\end{table}

\subsubsection{Label Distillation}
We evaluate the usefulness of our annotations by training on generated labels (i.e. label distillation) and compare the performance with a model of the same architecture (all-mpnet-base-v2 with 100M parameters) but trained on the full training dataset. In both settings, only the classification head is trained. The label distillation experiments use labels generated by Truncated RAC (L+D), which cover a portion of the original training datasets with various level of noise.

Table~\ref{table:distile_fft_f1} shows the results. We see that even with partial, noisy training data, the distilled model still shows generalization and achieves good performance on datasets with few categories. However, we note that there is a significant gap compared to the fine-tuned model for the more challenging datasets. Interestingly, training on noisy labels surpass the performance of Full RAC and even approaches Truncated RAC even though in distilled cases, the model predicts all test cases and thus has 100\% coverage.  

%While the performance of the distilled models falls below the performance of a classifier trained on the full gold datasets, AnnotateLLM only utilises a small pre-trained LLM and the information contained within a label schema. This demonstrates the viability of using Mistral-7B and written information relating to each category of interest (labels and descriptions) for the purpose of first-pass annotation and then distilling the information into a lightweight classifier. However, as also highlighted by~\citet{pangakis2023automated}, validation is still necessary to ensure the creation of high quality datasets.

\section{Results on internal datasets}
\begin{table*}[t!]
  \centering
\begin{tabular}{lllllll}
 & \multicolumn{3}{c}{1 step} & \multicolumn{3}{c}{5 step} \\
 Inferences & Micro & Macro &  Cov. & Micro & Macro &  Cov. \\
 \hline
 1 & 74.2 & 67.2 & 47.4 & 61.5 & 56.7 & 70.7 \\
 3 & 79.0 & 72.0 & 40.1 & 64.9 & 58.0 & 61.0 \\
 5 & 82.2 & 74.7 & 35.5 & 68.5 & 61.4 & 54.1 \\
\end{tabular}
  \caption{Truncated RAC experiments on internal dataset. }
  \label{table:internal_results}
\end{table*}
We now present our studies on our internal consumer banking dataset. Our dataset shares similar characteristics with the Banking77 dataset and contains 61 categories with detailed category descriptions. The business has relatively abundant data in this case, thus mainly requires an auto labelling tool with high label quality as long as the tool is able to annotate a significant percentage of cases. 

Based on our studies on public datasets, we use our best settings which runs RAC at maximally 5 steps with detailed label description. We cover 70.7\% of the cases with 61.5\% Micro-F1. However, the generated label quality does not meet the business requirements. To improve the label quality, we 1) further leverage the truncated tradeoffs by selecting only the most promising label as input to the LLM, together with the corresponding label description 2) apply self consistency~\cite{naik2023diversity} where we run $n$ inferences and only accept the answer in which all inferences agree with one another. The detailed results are shown in Table~\ref{table:internal_results} where the settings can trade-off between 60 F1 score to 80 F1 score; further improvement on label quality needs more powerful open-source LLMs, switching to few-shot cases with some manual efforts or more innovation upon the current label generation process.\footnote{We notice that RAC can be applied to multilabel business cases by considering $n$ retrieved cases and return all accepted labels.} 
\begin{table}[t!]
  \centering
\begin{tabular}{lllll}
  & \multicolumn{2}{c}{Distilled model} & \multicolumn{2}{c}{Finetuned model}\\
 Dataset &   Micro & Macro  &     Micro & Macro\\
 \hline
  AGNews   &  84.1 & 83.9 & \textbf{90.6} & \textbf{90.6}   \\
  Banking77 & 71.2 & 69.6 & \textbf{93.5} & \textbf{93.5}  \\
 DBpedia & 90.1 & 89.8 &  \textbf{98.5} & \textbf{98.5} \\ 
 Amazon  & 68.1 & 61.2 &  \textbf{86.2} & \textbf{83.1} \\
\end{tabular}

  \caption{Comparison between a label distillation model and a finetuned model with the same architecture but differs in the training datasets.}
  \label{table:distile_fft_f1}
\end{table}

\section{Related Work}
%LLM for annotations, LLM for classification tasks. 
%Litterature for retrieval is very different, this is referenced later on (RAG litterature).  

\label{section:related_work}
%Typical end-to-end annotation projects, such as the one described by~\citet{tseng2020best}, require a coordinated effort between stakeholders (the business, data scientists, subject matter experts, etc.) to successfully move from ideation to the delivery of a high-quality labelled dataset. Each stage must be approached in such a way that minimises the need to revisit earlier work and reduces the burden on annotators, thus helping to avoid the high costs that are commonly associated with annotation projects~\citep{grimmer2013text}. Due to its high cost and lack of scalability, within this work we focus on improvements to, and automation of, the annotation stage~\citep{carrell2016juice}.

It has been extensively shown that LLMs is a promising technique to automatically label data, reducing significantly annotation costs. \citet{wang-etal-2021-want-reduce} are arguably one of the first along this line, with GPT-3 as their LLM. With ChatGPT and GPT-4~\citep{openai2023gpt4}, the LLM labeling techniques give more accurate labels. For example, ~\citet{gilardi2023chatgpt} demonstrate the zero-shot abilities of ChatGPT (gpt-3.5-turbo), and find superior accuracy versus MTurk workers on 4 out of 5 datasets relating to political science; auto labelling benchmarks emerge enabling studying and comparing in a systematic way~\footnote{https://github.com/refuel-ai/autolabel}. We follow these works but also take into account real-life constraints including privacy concerns and computing resource provision concerns, leading us to consider ``small'' LLMs for these tasks.  

To improve the labelling quality for our considered LLMs, we mainly follow~\citep{fonseca2023can} to integrate label schema information into LLMs. While we show it improves performance on binary classification in preliminary results, we also show that it is non trivial to extend to a challenging multi-class classification setting. We propose RAC, which casts multi-class classification into a series of binary classification ones. Adding a retrieval step certainly draws inspiration from RAG~\citep{DBLP:journals/corr/abs-2005-11401} for classification settings. The closest literature we found is \citep{sun2023text} which uses a kNN retrieval step for classification task. RAC is different from \citep{sun2023text} in the considered settings. We consider zero-shot settings thus only retrieve label schema information, while \citep{sun2023text} considers few-shot settings and retrieve training data from its kNN module; secondly, the LLM in RAC considers ranked list candidates in binary classification in an iterative approach, which enables a balance between label quality and coverage, as we show in the Truncated RAC experiments.

\section{Conclusion}
\label{section:conclusion}
In this work, we study label generation performance under privacy and resource provisioning. We propose Retrieval Augmented Classification (RAC) to integrate label information into LLM based label generation. RAC reformulates challenging multiclass classification into a sequential of binary classifications starting from the most promising label to the least promising ones. We show that the framework allows integrating label schema information, which improves the generated label quality. Furthermore, by focusing solely on promising labels, RAC allows efficiently trade-offs between generated label quality and data coverage. When abstaining, the existing techniques generate labels of high quality for simple datasets that can be distilled efficiently even compared to fine-tuning cases, although there is still a performance gap for more challenging datasets such as our internal dataset.

%In this work, we have presented our approach to efficiently leverage small LLMs in zero-shot scenarios. We find that using label schema information, specifically category labels and descriptions, with self-consistency in a RAC pipeline, allows us to, in part, address the need for in-house automatic annotation. We investigated a multi-agent framework to create and refine annotation guidelines, finding that zero-shot performance is improved on the Banking77 dataset. However, further work is required to see consistent gains across a wider range of data. 

\bibliographystyle{plain}
\bibliography{references}

\appendix
\section{Llama 2 7B preliminary results}
\begin{table}[H]
  \centering
  \begin{tabular}{llllll}
                   & & L & L+D \\
                  \hline
 & No CoT  & \textbf{71.7} & \textbf{72.3}\\
 & CoT  & 58.4 & 58.6 \\
\end{tabular}
  \caption{Macro-averaged zero-shot classification accuracy across datasets as the amount of label schema information in the prompt is varied. Category labels (L) and a combination of both (L+D) are provided. A comparison is included between Chain-of-Thought (CoT) and non-CoT prompts.}
  \label{table:appendix:llama_results}
\end{table}

\newpage
\section{Self consistency}
\label{section:appendix:self_consistency_experiments}

\begin{itemize}
    \item Majority voting - Simple majority voting is applied across multiple binary predictions.
    \item Meta-reasoning - The rationales from multiple predictions are used within a final prompt that the LLM can reason over to make a final prediction.
    \item Weak supervision - A weakly supervision model~\citep{ratner2017snorkel} is applied to the predictions from the LLM~\citep{arora2022ask}. 
\end{itemize}

\begin{table}[H]
  \centering
  \begin{tabular}{llll}
 & \multicolumn{3}{c}{Prompts including} \\
 Method & No CoT      & CoT & Both\\
 \hline
 BS & 0.0 &  +0.4 &  +0.2\\
 MV & \textbf{+1.6} & \textbf{+2.3} & \textbf{+2.6}\\
 WS &  +1.3 &  +1.5 & +2.1 \\
 MR & +1.5 &  +0.7 & N/A    
\end{tabular}
  \caption{Relative macro-averaged zero-shot accuracy of Mistral-7B using category labels and descriptions. Self-consistency methods MV (majority voting), WS (weak supervision), and MR (meta-reasoning) are compared to a baseline (BS) of the average performance across the permutations of labels and descriptions.  We explore the effect of 5 inferences with or without CoT prompting, and the impact of using both configurations simultaneously.}
  \label{table:self_consistency}
\end{table}

\section{Prompt Templates}
\label{section:appendix:templates}
The following figures contain examples of the prompt templates used throughout the experiments.

\begin{figure*}[!ht]
\centering
\begin{center}
\fbox{
\fbox{
\parbox{5.5in}{
Your primary goal is to create a high-quality category name 

from \{\{ domainDataDescription \}\} samples provided by the user. 
\\\\
All samples come from the same category, 

you need to create an appropriate name for the category.
\\\\
The name should be no more than 10 words. 

The name should be a concise and clear label for the category. 

It can be either verb phrases or noun phrases, whichever is more appropriate.
\\\\
Consider the content in all samples, 

the name must be applicable to all samples.

Only output the category name, include no other text.

Samples:

...

Created category name:
}}}
\end{center}
\caption{Prompt used to generate a category label using the text from samples belonging to the same class. Prompt design follows~\citet{shah2023using}. Jinja2 templating is used to insert data into the prompt.}
\label{figure:prompt_label_generation}
\end{figure*}

\begin{figure*}[!ht]
\centering
\begin{center}
\fbox{
\fbox{
\parbox{5.5in}{
Now your primary goal is to generate an high-quality category description 

from the given \{\{ domainDataDescription \}\} samples.

The description of the category should be no more than 50 words 

and give enough information to identify the category in the samples.
\\\\
Consider the content in all samples, 

the description must be applicable to all samples.

It should be thorough and detailed, highlighting nuances.

It should describe and define the category.
\\\\
Only output the description, 

include no other text.

Created category description:
}}}
\end{center}
\caption{Prompt used to generate a category description using the text from samples belonging to the same class. Prompt design follows~\citet{shah2023using}. Jinja2 templating is used to insert data into the prompt.}
\label{figure:prompt_description_generation}
\end{figure*}

\begin{figure*}[!ht]
\centering
\begin{center}
\fbox{
\fbox{
\parbox{5.5in}{
You are an advanced AI model designed to annotate and classify data samples 
\\
using information provided by the user.
\\\\
Think carefully before giving your final answer.
\\\\
Write out your reasoning step-by-step to be sure you get the right answer!
\\\\
Help me classify a \{\{ domainDataDescription \}\}  \\
sample against the following category label and description.
\\\\
Category label:  \{\{ categoryLabel \}\}
Category description:  \{\{ categoryDescription \}\}
\\\\
Sample: {{ text }}
\\\\
The category must directly relate to the sample, 
\\
otherwise they are not related.
\\\\
Does the sample directly belong to the category described in the label and description?
Choices, yes or no? Answer:
}}}
\end{center}
\caption{Prompt used to classify a sample against a category described by the label and description using No-COT configuration. Jinja2 templating is used to insert data into the prompt.}
\label{figure:prompt_label_classification_no_cot}
\end{figure*}

\begin{figure*}[!ht]
\centering
\begin{center}
\fbox{
\fbox{
\parbox{5.5in}{
You are an advanced AI model designed to annotate and classify data samples 

using information provided by the user.
\\\\
Think carefully before giving your final answer.
\\\\
Write out your reasoning step-by-step to be sure you get the right answer!
\\\\
Help me classify a \{\{ domainDataDescription \}\} 

sample against the following category label and description.
\\\\
Category label: \{\{ categoryLabel \}\}

Category description: \{\{ categoryDescription \}\}
\\\\
Sample: {{ text }}
\\\\
The category must directly relate to the sample, 

otherwise they are not related.
\\\\
Does the sample directly belong to the category described 

in the label and description, explain your answer step-by-step?
}}}
\end{center}
\caption{Prompt used to classify a sample against a category described by the label and description using COT configuration. We encourage a rationale to be produced by including ``explain your answer step-by-step''. Jinja2 templating is used to insert data into the prompt.}
\label{figure:prompt_label_classification_cot}
\end{figure*}

%\appendix
%\section{System Diagrams}
%\label{section:appendix}

%Shown in Figure~\ref{figure:rac_diagram} is a high-level system diagram of the RAC pipeline for multi-class classification. In Figure~\ref{figure:multi_agent_diagram} the multi-agent framework for generating and improving annotation guidelines is outlined.

%\begin{figure*}[h!]
%    \centering
%    \includegraphics[width=\textwidth]{RAC_diagram.png}
%    \caption{Diagram of Retrieval Augmented Classification (RAC).}
%    \label{figure:rac_diagram}
%\end{figure*}

%\begin{figure*}[h!]
%    \centering
%    \includegraphics[width=\textwidth]{Automatic Guideline Improvements - Revised.png}
%    \caption{Diagram of the multi-agent system for the automatic refinement of annotation guidelines.}
%    \label{figure:multi_agent_diagram}
%\end{figure*}

\end{document}